\documentclass[12pt]{article}

\usepackage{graphicx}
\usepackage{mdframed}
\usepackage{hyperref}
\usepackage[a4paper, left=0.8in, right=0.8in, top=1in, bottom=1in]{geometry}
\usepackage{setspace}
\usepackage[numbers,sort&compress]{natbib}

\usepackage{xcolor}
\usepackage{microtype}
\usepackage{subcaption}
\usepackage{float}
\usepackage{booktabs}
\usepackage{amsmath, amssymb, mathtools, amsthm}
\usepackage{multirow}
\usepackage[capitalize,noabbrev]{cleveref}
\usepackage{listings}
\usepackage[most]{tcolorbox}

\definecolor{gsoboxblue}{HTML}{1f78b4}

\definecolor{diffadd}{RGB}{34,134,58}
\definecolor{diffrem}{RGB}{203,36,49}
\definecolor{diffhunk}{RGB}{111,66,193}
\definecolor{difffile}{RGB}{0,92,197}

\newtcolorbox{gsoboxblue}[1][]{
    enhanced,
    colback=white,
    colframe=gsoboxblue,
    coltitle=white,
    fonttitle=\bfseries,
    attach boxed title to top left={yshift=-2mm, xshift=0mm},
    boxed title style={
      colback=gsoboxblue,
      sharp corners,
      boxrule=0pt,
    },
    sharp corners,
    boxrule=0.5pt,
    left=6pt,
    right=6pt,
    top=4pt,
    bottom=4pt,
    breakable,
    #1
}

\theoremstyle{plain}

\theoremstyle{definition}

\theoremstyle{remark}

\usepackage[textsize=tiny]{todonotes}

\title{%
  \includegraphics[width=0.5\linewidth]{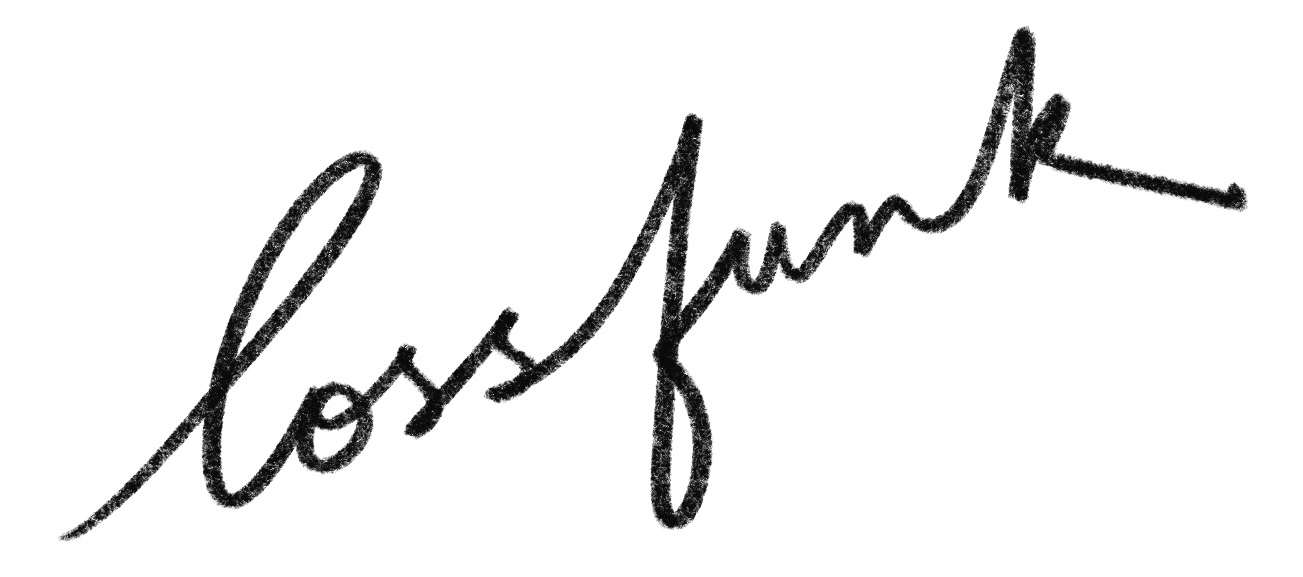}\\[1cm]
  ISO-Bench: Can Coding Agents Optimize Real-World Inference Workloads?
}
\author{%
  \textbf{Ayush Nangia}\quad \textbf{Shikhar Mishra}\quad \textbf{Aman Gokrani}\quad \textbf{Paras Chopra}\\[6pt]
  \texttt{\{ayush.nangia, shikhar.mishra, paras\}@lossfunk.com}
}
\date{February 2026}

\begin{document}

\maketitle

\begin{abstract}
We introduce ISO-Bench, a benchmark for coding agents to test their capabilities on real-world inference optimization tasks. These tasks were taken from vLLM and SGLang, two of the most popular LLM serving frameworks. Each task provides an agent with a codebase and bottleneck description, whereby the agent must produce an optimization patch evaluated against expert human solutions. We curated 54 tasks from merged pull requests with measurable performance improvements. While existing benchmarks heavily use runtime-based metrics, such approaches can be gamed to pass tests without capturing the actual intent of the code changes. Therefore, we combine both hard (execution-based) and soft (LLM-based) metrics to show that both are necessary for complete evaluation. While evaluating both closed and open-source coding agents, we find no single agent dominates across codebases.
Surprisingly, agents often identify correct bottlenecks but fail to execute working solutions. We also show that agents with identical underlying models differ substantially, suggesting scaffolding is as important as the model.
\end{abstract}

\section{Introduction}
LLM inference engines have become essential for deploying large language models at scale. Systems like vLLM \citep{kwon2023vllm} and SGLang \citep{zheng2023sglang} handle production workloads in industry and research, achieving high throughput through systems-level optimizations. Methods such as PagedAttention~\citep{kwon2023vllm} and FlashAttention~\citep{dao2022flashattention} required extensive work and in-depth knowledge in memory management, kernel development, and scheduling. The need for optimization is growing as more models are released and model architectures continuously evolve.
LLM-based coding agents have become powerful tools for software engineering, capable of finding bugs and generating patches across codebases. Systems like SWE-Agent~\citep{yang2024sweagent} and OpenHands~\citep{wang2025openhands} perform well on benchmarks like SWE-bench~\citep{jimenez2024swebench}.
However, recent benchmarks suggest these agents still struggle with optimization tasks. KernelBench~\citep{ouyang2025kernelbench} finds that frontier models match GPU kernel baselines in under 20\% of cases, GSO~\citep{shetty2025gso} reports success rates below 5\% on repository-level tasks, and SWE-Perf~\citep{he2025sweperf} observes large gaps between agent and expert solutions.
These benchmarks measure whether agents succeed, but not why they fail.
GSO takes a step further by combining execution metrics with LLM-as-a-Judge, but a key question remains: when agents fail, do they misunderstand the problem, or do they understand it but struggle to implement the solution?

In this work, we present ISO-Bench, a benchmark of 54 optimization tasks from vLLM and SGLang.
Beyond measuring throughput gains, we evaluate whether agents target the correct bottleneck and use appropriate strategies.

Our contributions are mentioned as follows:
\begin{enumerate}
    \item \textbf{Benchmarking Tasks}: 54 optimization tasks extracted from real commits in vLLM and SGLang. Each task includes a repository snapshot, throughput and latency benchmarks, and correctness tests.
    \item \textbf{Dual evaluation framework}: Hard and soft metrics are introduced that distinguish true successes from lucky wins, showing that traditional metrics might overestimate agent capabilities.
    \item \textbf{Behavioral insights}:  Identification of an understanding-execution gap as a primary failure mode and showing that agent performance varies substantially across codebases.
\end{enumerate}

\section{Related Work}
\label{sec:related}

\begin{figure}[t]
\centering
\includegraphics[width=\textwidth]{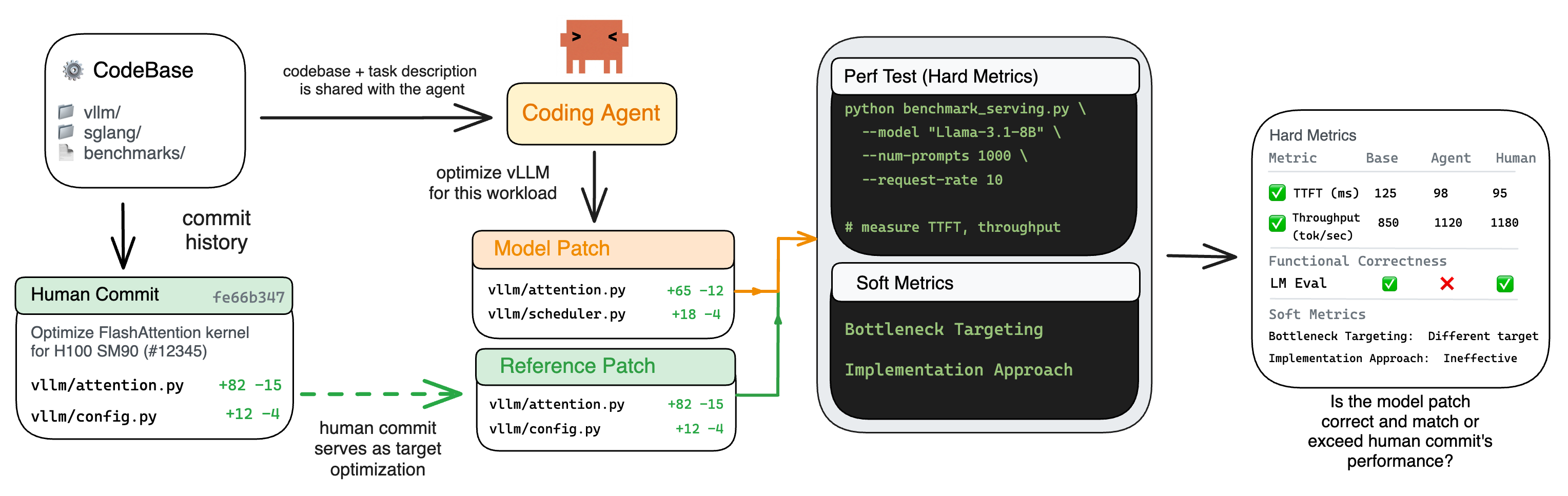}
\caption{\textbf{ISO-Bench evaluation pipeline.} Given a codebase and task description, a coding agent produces an optimization patch. We compare this
  patch against the human commit using hard metrics (TTFT, throughput) and soft metrics (bottleneck targeting, implementation approach). Hard metrics
  measure performance improvement; soft metrics assess whether the agent targeted the correct code.} \label{fig:pipeline}
\end{figure}

We review earlier works on evaluating and building LLM systems for code generation, focusing on: (i) benchmarks evolving from correctness to efficiency, (ii) coding agent architectures, and (iii) LLM-based evaluation methods.

\paragraph{Correctness-driven benchmarks:}Early benchmarks for code generation focused on measuring functional correctness for standalone functions. HumanEval~\citep{chen2021evaluating} introduced 164 hand-crafted Python problems with unit tests and popularized the pass@$k$ metric, which measures whether at least one of $k$ sampled solutions passes all tests. However, such function-level benchmarks evaluate code generation separate from the complexities of real software development.
Repository-scale benchmarks are more realistic as agents must navigate the full codebase, find the relevant code, and edit multiple files. SWE-bench~\citep{jimenez2024swebench} constructs tasks from real GitHub issues in popular Python repositories, requiring agents to pass executable tests. SWE-bench Verified~\citep{chowdhury2024swebenchverified} filters for reproducible evaluation and has become a standard target for coding agents.
\paragraph{Efficiency-driven benchmarks:}
Optimizing for performance is a different problem: agents must find the bottleneck, and success is measured by actual speedup rather than just correctness.
At the kernel level, KernelBench \citep{ouyang2025kernelbench} evaluates LLMs on generating efficient GPU kernels for 250 PyTorch ML workloads, introducing the \texttt{fast\_p} metric to count solutions that are both correct and achieve greater than $p\times $
 speedup over a PyTorch baseline; even strong models succeed on fewer than 20\% of tasks. Complementing this,
 TritonBench \citep{li2025tritonbench} offers two evaluation suites: \textit{TritonBench-G} (GitHub-sourced operators) and \textit{TritonBench-T} (PyTorch-aligned tasks), profiling Triton code against reference implementations and reporting both correctness and GPU efficiency.
\\\\
At the repository level, several recent benchmarks investigate if agents can optimize real-world codebases. SWE-Perf \citep{he2025sweperf} constructs tasks from
pull requests that improve performance, ensuring that patches are applied successfully, pass tests, and produce quantifiable speedups.  GSO \citep{shetty2025gso} evaluates agent patches against human expert commits rather than fixed thresholds, while SWE-fficiency \citep{ma2025swefficiencylanguagemodelsoptimize} scales this methodology across a broader set of Python libraries, reporting how close agents come to expert-level improvements. Across all three repository-level benchmarks, agents consistently struggle to identify bottlenecks and understand low-level performance.

\paragraph{Coding Agent Architectures:} Recent work has shown that agent scaffolding significantly impacts software engineering performance. SWE-Agent \citep{yang2024sweagent} demonstrated that agent-computer interfaces are important for better code manipulation, while OpenHands~\citep{wang2025openhands} complements this by providing an open platform for designing and evaluating coding agents using standardized tools and benchmarks. TRAE-Agent~\citep{gao2025traeagent} targets repository-level tasks by generating candidate patches, pruning them, and selecting a final solution. In contrast, commercial systems such as Claude Code~\citep{anthropic_claude_code} and Codex CLI~\citep{openai_codex_cli} couple the model with proprietary scaffolding, which makes it difficult to separate gains from the underlying LLM versus the agent architecture.

\paragraph{LLM-as-a-Judge for Code Evaluation:} Using LLMs to evaluate code has gained traction as a scalable alternative to test-based evaluation. \citet{zheng2023judging} showed that strong LLM judges can match human preferences with over 80\% agreement. ICE-Score \citep{zhuo2024ice} applies this to code by guiding LLMs through step-by-step assessment, while CodeJudge \citep{tong2024codejudge} enhances accuracy by asking LLMs to think about error categories before scoring. Despite their promises, LLM judges suffer from biases such as favoring lengthier outputs and preferring their own generations, while code evaluation brings additional biases like susceptibility to misleading remarks \citep{moon2025codebias}.

\section{ISO-Bench}
\label{sec:iso-bench}
Existing benchmarks evaluate either standalone kernel generation (KernelBench, TritonBench) or general repository optimization (GSO, SWE-Perf, SWE-fficiency), but none target the specific challenges of GPU-based inference serving systems.
\\\\
ISO-Bench fills this gap by focusing on execution-based metrics for real-world, GPU-based inference optimization workloads.
Figure \ref{fig:pipeline} shows the end-to-end ISO-Bench evaluation pipeline, from task inputs and agent-generated patches to hard/soft metrics and correctness validation.

In this section, we describe how tasks are formulated, how the benchmark was constructed, and how we evaluate agent performance.
\subsection{Task Formulation}
\label{sec:task-formulation}

Each ISO-Bench task presents an agent with two inputs: (i)~the repository at a pre-optimization commit state (ii)~a task description explaining which performance bottleneck to address without revealing the solution. The agent must produce an optimization patch that improves performance on the specified benchmark. We evaluate agent patches against human expert solutions from the original pull requests.

\subsection{Benchmark Construction}
\label{sec:benchmark-construction}

We collect optimization tasks from two production ML inference engines: \textbf{vLLM}~\citep{kwon2023vllm} and \textbf{SGLang}~\citep{zheng2023sglang}. We selected vLLM and SGLang primarily because they are widely used inference engines. This enables a realistic evaluation of agent performance on production-grade inference optimization tasks.

\paragraph{Stage 1: Commit Extraction:} We use a GSO-inspired pipeline to identify performance-related commits through keyword filtering (terms like \texttt{optim}, \texttt{speed}, \texttt{latency}, \texttt{memory}) followed by LLM-based classification to only keep commits focused on GPU-based inference optimization (see Appendix~\ref{app:task-collection} for filtering statistics).

\paragraph{Stage 2: Manual Curation:} We manually review each candidate commit to verify that (i)~the optimization is reproducible (ii)~the commit represents a genuine optimization rather than a refactoring or bug fix. This curation step filters out false positives from the automated pipeline.

\paragraph{Stage 3: PR Analysis:} For each filtered commit, we fetch the associated pull request to extract the benchmarking model, evaluation commands, and performance claims from the PR discussion. This metadata serves two purposes: it provides the benchmark commands we use during evaluation, and it helps manual curation by helping us verify whether a commit represents a legitimate optimization.

This pipeline produces \textbf{54 benchmark instances}: 39 from vLLM and 15 from SGLang, each with verified performance benchmarks, model specifications, and evaluation commands.

\subsection{Evaluation Metrics}
\label{sec:evaluation-metrics}
We evaluate agent performance using two types of metrics. Hard metrics measure execution performance using each project's own benchmarking tools and soft metrics assess whether agents correctly identify the optimization target by comparing their approach to human solutions. Combining both allows us to distinguish genuine optimization capability from accidental improvements.

\subsubsection{Hard Metrics}
\label{sec:hard-metrics}
We measure agent optimizations using the same benchmarks that developers used in the original pull requests. We track Time to First Token (TTFT) and throughput, comparing agent performance against the human baseline and classifying results according to Table~\ref{tab:hard-metric-categories}:
\begin{equation}
\Delta_{\text{TTFT}} = \frac{\text{TTFT}_h - \text{TTFT}_a}{\text{TTFT}_h} \times 100
\label{eq:ttft}
\end{equation}
\begin{equation}
\Delta_{\text{throughput}} = \frac{\text{Throughput}_a - \text{Throughput}_h}{\text{Throughput}_h} \times 100
\label{eq:throughput}
\end{equation}

\begin{table}[t]
\caption{Classification of hard metrics based on performance delta.}
\label{tab:hard-metric-categories}
\centering
\small
\begin{tabular}{@{}ll@{}}
\toprule
\textbf{Category} & \textbf{Criteria} \\
\midrule
Beats & Agent improves on human by $>5\%$ \\
Similar & Agent within $\pm 5\%$ of human \\
Worse & Agent degrades by $>5\%$ \\
Failed & Patch causes benchmarking error \\
\bottomrule
\end{tabular}
\end{table}

We use a 5\% threshold to account for measurement noise. For serving-based benchmarks, we measure latency using $\Delta_{\text{TTFT}}$. When TTFT is not produced, we use $\Delta_{\text{throughput}}$ for standalone benchmarks.

\subsubsection{Soft Metrics}
\label{sec:soft-metrics}
Hard metrics alone cannot distinguish genuine optimization capability from accidental improvements. Agents may achieve performance gains through changes unrelated to the actual optimization target. To address this, we introduce soft metrics, which uses LLM-as-a-Judge (Gemini-3-Flash-Preview~\citep{deepmind2025gemini3flash_modelcard}) to compare agent patches against human solutions. We consider two dimensions as shown in Table~\ref{tab:soft-metric-categories}.

\begin{table}[t]
\caption{Soft metric categories.}
\label{tab:soft-metric-categories}
\centering
\small
\begin{tabular}{@{}ll@{}}
\toprule
\multicolumn{2}{l}{\textbf{Bottleneck Targeting}} \\
\midrule
Same target & Identical code locations as human \\
Related target & Same module or subsystem \\
Different target & Unrelated code areas \\
No optimization & No performance-relevant changes \\
\midrule
\multicolumn{2}{l}{\textbf{Implementation Approach}} \\
\midrule
Similar approach & Same technique as human \\
Valid alternative & Different but sound \\
Partial solution & Subset of required changes \\
Ineffective & Fails to address bottleneck \\
\bottomrule
\end{tabular}
\end{table}

\subsubsection{Quadrant Framework}
\label{sec:quadrant-framework}

We combine hard and soft metrics to classify each optimization attempt into one of four quadrants, as shown in
Figure~\ref{fig:quadrant-framework}. The framework maps hard metrics performance classification
(Table~\ref{tab:hard-metric-categories}) against soft metrics categories (Table~\ref{tab:soft-metric-categories}):

\begin{figure}[t]
\centering
\includegraphics[width=0.85\textwidth]{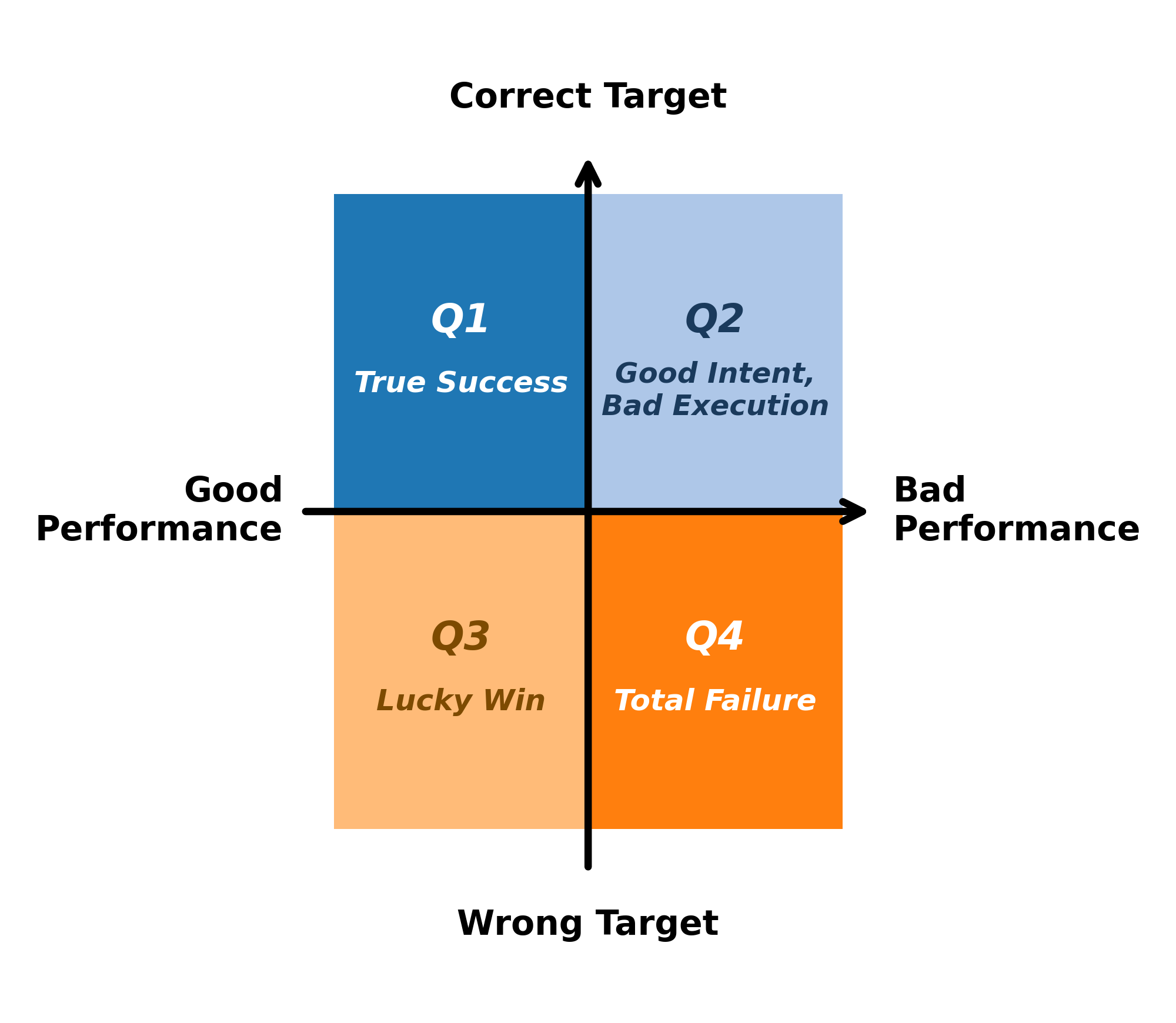}
\caption{Quadrant framework for evaluating optimization attempts. The horizontal axis shows performance (good: beats or similar; bad: worse or failed). The vertical axis shows whether the agent targeted the correct bottleneck (correct: same or related target; wrong: different target or no optimization). Q1 True Success: correct target, good performance. Q2 Good Intent: correct target, bad performance. Q3 Lucky Win: wrong target, good performance. Q4 Total Failure: wrong target, bad performance.}
\label{fig:quadrant-framework}
\end{figure}

\begin{itemize}
\item \textbf{Q1 (True Success):} Agent targets the correct bottleneck (Same or Related target) and achieves competitive performance (Beats or Similar).

\item \textbf{Q2 (Good Intent, Bad Execution):} Agent targets the correct bottleneck (Same or Related target) but fails to achieve performance gains (Worse or Failed). The agent understood the problem but could not implement a working solution.

\item \textbf{Q3 (Lucky Win):} Agent achieves good performance (Beats or Similar) despite targeting the wrong code (Different target or No optimization). Without soft metrics, these would be classified as success.

\item \textbf{Q4 (Complete Failure):} Agent targets the wrong code (Different target or No optimization)
and fails to achieve performance gains (Worse or Failed).
\end{itemize}

This framework gives two success measurements:
\begin{itemize}
  \item \textbf{True Success} = Q1 (requires both correct targeting and performance improvement)
  \item \textbf{Hard Success} = Q1 + Q3 (based on hard metrics only)
\end{itemize}

Only True Success measures actual optimization ability.

\subsubsection{Functional Correctness}
\label{sec:functional-correctness}
The quadrant framework identifies Hard Success cases (Q1 + Q3) where agents achieve performance improvements. However, speedup alone does not guarantee correctness. An agent might achieve faster execution by changing model behavior in ways that produce incorrect outputs.

We validate functional correctness for all Hard Success cases using the LM Evaluation Harness~\citep{eval-harness}. For each task, we use the evaluation benchmarks specified in the original PR and measure accuracy for the unoptimized baseline and the agent patch. If accuracy remains consistent, the optimization preserves correctness. If the agent patch degrades accuracy, the optimization introduced functional errors despite achieving speedup.

This validation is especially important for Q3 (Lucky Win) cases. These agents achieve speedup without targeting the correct bottleneck, so their improvements may come from changes that alter model behavior rather than genuine optimization.

\section{Experimental Setup}
\label{sec:experimental-setup}

\begin{figure}[t]
    \centering
    \begin{minipage}[t]{0.48\textwidth}
        \centering
        \includegraphics[width=\textwidth]{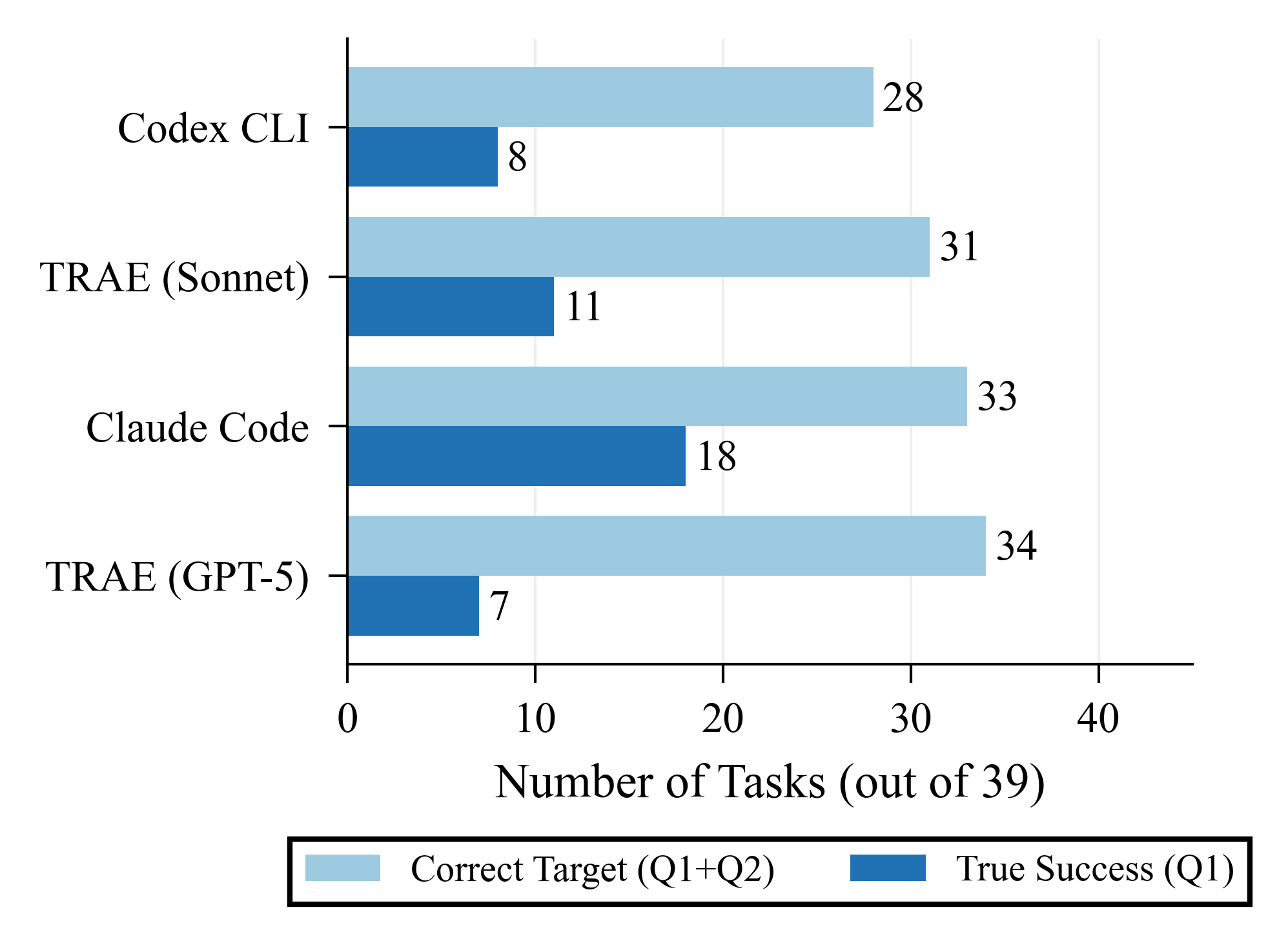}
        \caption{Good Intent vs Bad Execution on vLLM (39 tasks). Light bars show correct target identification
        (Q1+Q2). Dark bars show True Success (Q1). The gap represents Q2 failures.}
        \label{fig:good-intent-vllm}
    \end{minipage}
    \hfill
    \begin{minipage}[t]{0.48\textwidth}
        \centering
        \includegraphics[width=\textwidth]{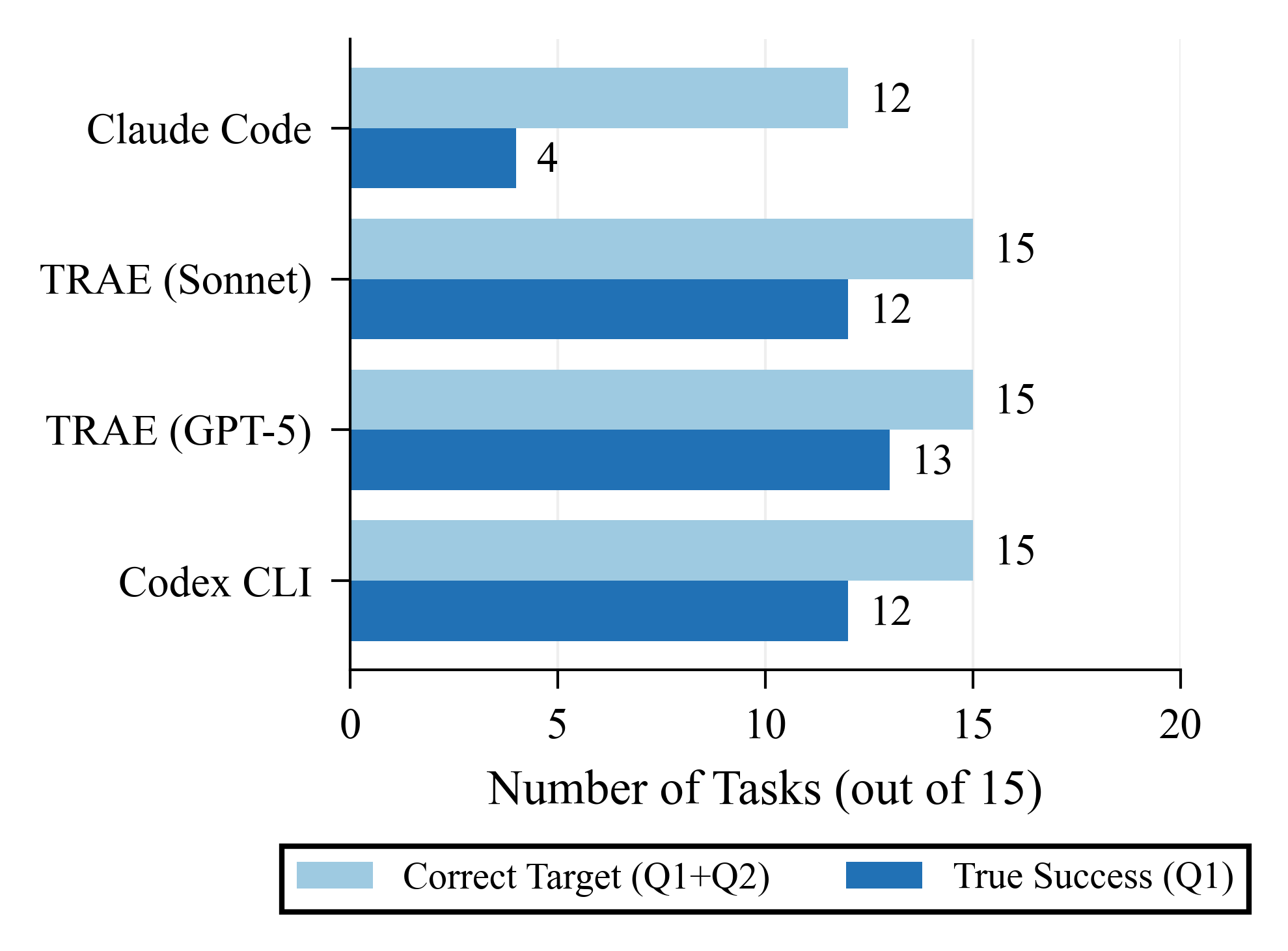}
        \caption{Good Intent vs Bad Execution on SGLang (15 tasks). Light bars show correct target identification
        (Q1+Q2). Dark bars show True Success (Q1). The gap represents Q2 failures.}
        \label{fig:good-intent-sglang}
    \end{minipage}
\end{figure}

This section describes the agent scaffolding we evaluate, the execution environment used, and the process for running tasks and scoring patches with hard and soft metrics.

\subsection{Agent Scaffolding}
\label{sec:agents}

We evaluate three coding agents with different models and scaffolding:
\begin{itemize}
  \item \textbf{Claude Code:} Anthropic's coding assistant designed for autonomous software engineering tasks using Claude Sonnet 4.5.
  \item \textbf{Codex CLI:} OpenAI's command-line coding interface, supporting iterative code editing, execution, and debugging from the terminal using GPT-5.
    \item \textbf{TRAE-Agent:} Modified version of ByteDance's open-source TRAE-Agent framework, evaluated with two underlying models (Claude Sonnet 4.5 and GPT-5). We refer to these as TRAE (Sonnet) and TRAE (GPT-5) respectively.
\end{itemize}
Full agent configurations are detailed in Appendix~\ref{app:task-details}.

\begin{figure}[t]
    \centering
    \begin{minipage}[t]{0.48\textwidth}
        \centering
        \includegraphics[width=\textwidth]
        {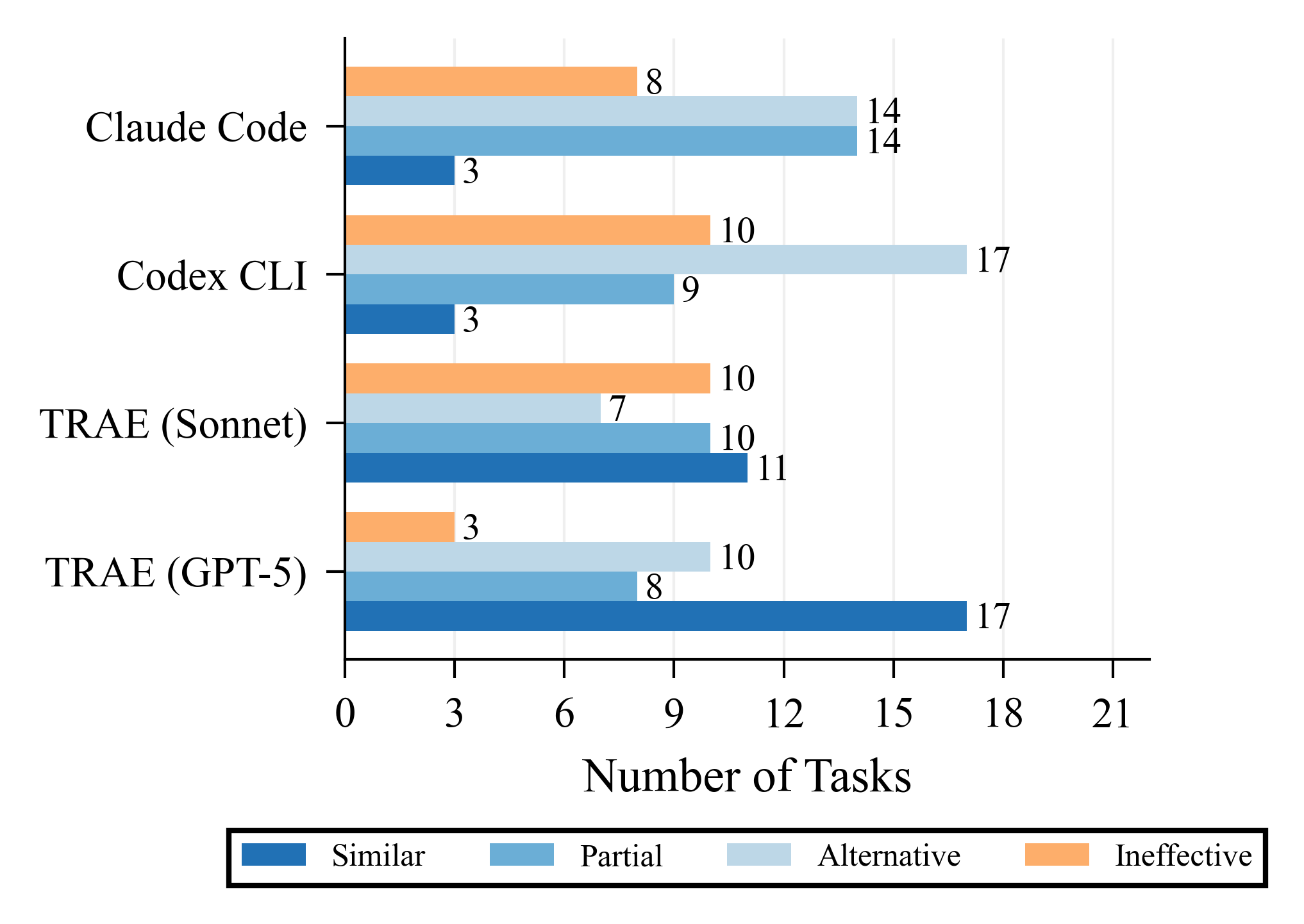}
        \caption{Approach distribution on vLLM (39 tasks).}
        \label{fig:approach-vllm}
    \end{minipage}
    \hfill
    \begin{minipage}[t]{0.48\textwidth}
        \centering
        \includegraphics[width=\textwidth]
        {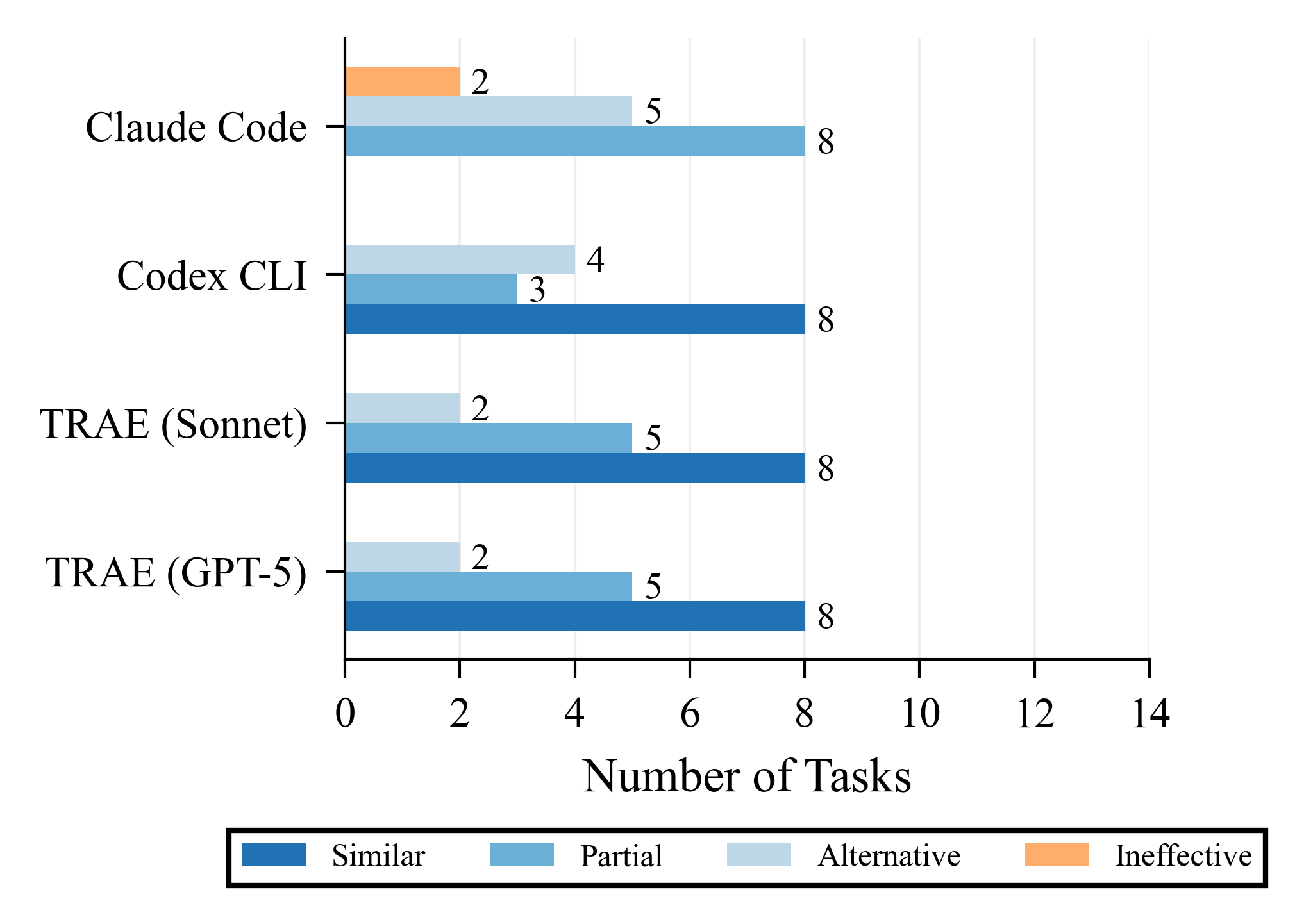}
        \caption{Approach distribution on SGLang (15 tasks).}
        \label{fig:approach-sglang}
    \end{minipage}
\end{figure}

\subsection{Execution Environment}
\label{sec:execution-environment}

Each agent operates on an isolated git worktree where it can freely explore the codebase, modify files, and commit changes. Agents have 120 minutes per task to work on their solution. During this time, they can run tests, check performance, and refine their code based on the results. All runs happen inside Docker containers to keep measurements consistent across experiments. We save all agent edits as git commits so we can analyze them later. Each agent receives a structured task prompt describing the optimization target without revealing the solution (see Appendix~\ref{app:task-details} for full environment specifications and the task prompt template).

\subsection{Evaluation Protocol}
\label{sec:evaluation-protocol}

After all agent runs complete, we evaluate patches in two stages. For hard metrics, we execute the benchmark commands on NVIDIA H100 GPUs, measuring TTFT and throughput against both the unoptimized baseline and the human solution. For soft metrics, we perform LLM-based (LLM-as-a-Judge) analysis comparing agent patches to human patches, assessing bottleneck targeting and implementation approach.

\section{Results}
We evaluate three coding agents on ISO-Bench across 54 optimization tasks (39 from vLLM, 15 from SGLang). This section presents our findings on how coding agents approach inference optimization challenges and why traditional metrics alone fail to capture their performance.

\subsection{Can Agents Optimize GPU Inference Code?}
The percentage of tasks where agents both identified the correct bottleneck and achieved measurable performance improvements is defined as the True Success rate, presented in Table~\ref{tab:true-success}. Appendix~\ref{app:flash-attention-case-study} provides a detailed Q1 example.

\begin{table}[H]
\centering
\caption{True Success rates (\%) by Project.}
\label{tab:true-success}
\begin{tabular}{lcc}
\toprule
\textbf{Agent} & \textbf{vLLM} & \textbf{SGLang} \\
\midrule
Claude Code & 46.2\% & 26.7\% \\
TRAE (Sonnet) & 28.2\% & 80.0\% \\
TRAE (GPT-5) & 17.9\% & 86.7\% \\
Codex CLI & 20.5\% & 80.0\% \\
\bottomrule
\end{tabular}
\end{table}

In our experiments, we study the two inference engines vLLM and SGLang separately. On vLLM, Claude Code gets 46.2\%, while the other agents get lower True Success rates. However, on SGLang, Claude Code gets 26.7\%, while the other agents get higher True Success rates, with TRAE (GPT-5) at 86.7\%. This difference in agent performance for these projects is further studied in Section~\ref{sec:generalize}.
\begin{table}[t]
    \centering
    \caption{Hard Success vs.\ True Success rates. The Gap column quantifies cases where agents achieved performance improvements without targeting the correct bottleneck. A gap of zero indicates all hard successes genuinely addressed the specified optimization target.}
    \label{tab:hard-vs-true}
    \begin{tabular}{@{}llccc@{}}
    \toprule
    \textbf{Project} & \textbf{Agent} & \textbf{Hard Success (\%)} & \textbf{True Success (\%)} & \textbf{Gap (\%)} \\
    \midrule
    \multirow{4}{*}{vLLM}
      & Claude Code   & 56.4 & 46.2 & 10.2 \\
      & Codex CLI     & 33.3 & 20.5 & 12.8 \\
      & TRAE (Sonnet) & 33.3 & 28.2 &  5.1 \\
      & TRAE (GPT-5)  & 20.5 & 17.9 &  2.6 \\
    \midrule
    \multirow{4}{*}{SGLang}
      & Claude Code   & 46.7 & 26.7 & 20.0 \\
      & Codex CLI     & 80.0 & 80.0 &  0.0 \\
      & TRAE (Sonnet) & 80.0 & 80.0 &  0.0 \\
      & TRAE (GPT-5)  & 86.7 & 86.7 &  0.0 \\
    \bottomrule
    \end{tabular}
\end{table}

\subsection{Do Hard Metrics Tell the Full Story?}

Each task in ISO-Bench specifies a particular bottleneck that agents must optimize. However, agents sometimes achieve measurable speedups by modifying code unrelated to the specified bottleneck. These are Lucky Wins (Q3) from Section~\ref{sec:quadrant-framework}, and the gap between Hard Success and True Success measures how often this occurs.

Table~\ref{tab:hard-vs-true} presents the complete breakdown across all agent-project combinations. Claude Code on SGLang achieves a Hard Success rate of 46.7\%, but under True Success this drops to 26.7\%, a gap of 20\%. Similar patterns emerge across other configurations, with gaps ranging from 2.6\% to 12.8\% on vLLM. These results show that hard metrics alone can overestimate agent capabilities. Soft metrics address this by checking whether agents actually modify the code regions specified in each task, making it possible to separate understanding from accidental improvements.

\begin{table}[b]
    \centering
    \caption{Distribution of outcomes across quadrants (Q1-Q4).}
    \label{tab:quadrant-distribution}
    \begin{tabular}{@{}llcccc@{}}
    \toprule
    \textbf{Project} & \textbf{Agent} & \textbf{Q1} & \textbf{Q2} & \textbf{Q3} & \textbf{Q4} \\
    \midrule
    \multirow{4}{*}{vLLM}
      & Claude Code   & 18 & 15 & 4 & 2 \\
      & Codex CLI     &  8 & 20 & 5 & 6 \\
      & TRAE (Sonnet) & 11 & 20 & 2 & 6 \\
      & TRAE (GPT-5)  &  7 & 27 & 1 & 4 \\
    \midrule
    \multirow{4}{*}{SGLang}
      & Claude Code   &  4 &  8 & 3 & 0 \\
      & Codex CLI     & 12 &  3 & 0 & 0 \\
      & TRAE (Sonnet) & 12 &  3 & 0 & 0 \\
      & TRAE (GPT-5)  & 13 &  2 & 0 & 0 \\
    \bottomrule
    \end{tabular}
\end{table}

\subsection{Why Do Agents Fail?}
\label{sec:what-limits}

 Prior benchmarks (e.g., GSO~\citep{shetty2025gso}, KernelBench~\citep{ouyang2025kernelbench}) often link poor agent performance to limited understanding of the optimization target. However, we observe that for inference optimization tasks, agents frequently identify correct bottleneck, yet fail to produce a working optimization. We classify these cases in Q2, Good Intent Bad Execution, where agents target the right optimization but fail to implement a working solution.

On vLLM, three of four agents have their highest count in Q2, as shown in Table \ref{tab:quadrant-distribution}. The gap between correct understanding and execution is illustrated in Figure \ref{fig:good-intent-vllm}. TRAE (GPT-5) has the highest understanding-to-execution gap. Other agents follow similar patterns, except for Claude Code which has the strongest execution capability. Appendix~\ref{app:qwen3-case-study} provides a detailed Q2 example where the agent identified the correct target but failed to implement a working solution. See Appendix~\ref{app:prefix-caching-case-study} for a Q4 complete failure example.

On SGLang, the gap narrows considerably; see Figure \ref{fig:good-intent-sglang}. All agents except Claude Code identify the correct target in all 15 tasks. Not only that, these agents also demonstrate much stronger execution. In contrast, Claude Code struggles on this codebase: despite identifying 12 correct targets, it fails to execute on 8 of them.

\subsection{Does Performance Generalize Across Codebases?}
\label{sec:generalize}

Agent rankings shift between vLLM and SGLang, with the top performer on one codebase becoming a bottom
performer on the other. To understand this switch, we examine each agent's optimization strategy relative to the human reference fix.

Figure~\ref{fig:approach-vllm} illustrates approach distribution on vLLM. No single strategy dominates, and matching the human approach does not guarantee success. TRAE (GPT-5) matches the human approach most frequently yet achieves the lowest True Success. In contrast, Claude Code rarely matches the reference, favoring Partial and Alternative approaches, yet achieves the highest True Success.

Figure \ref{fig:approach-sglang} highlights a different pattern on SGLang. All agents except Claude Code frequently adopt approaches similar to human approaches which results in higher success. Claude Code never adopts a Similar approach, a strategy that backfires on SGLang.

Each agent maintains its preferred strategy regardless of the project. Claude Code consistently favors alternative approaches, which succeed on vLLM but fail on SGLang. TRAE (GPT-5), TRAE (Sonnet) and Codex CLI consistently match the reference approach, which succeeds on SGLang but fails on vLLM. This suggests that single-codebase evaluations can overstate how well an agent generalizes to new repositories.

\subsection{Does Agent Scaffolding Matter?}
\label{sec:scaffolding}

TRAE (Sonnet) and Claude Code both use the same underlying model: Claude Sonnet 4.5. However, their performance differs notably across codebases as seen in Table~\ref{tab:hard-vs-true}. On vLLM, Claude Code achieves 46.2\% True Success while TRAE (Sonnet) reaches 28.2\%. On SGLang, TRAE (Sonnet) achieves 80.0\% while Claude Code reaches only 26.7\%.

The model is identical, but the agents differ in how they explore the codebase, decompose tasks, and decide when to stop iterating. These scaffolding choices produce the approach distributions shown in Figures~\ref{fig:approach-vllm} and~\ref{fig:approach-sglang}: Claude Code favors partial and alternative solutions, while TRAE (Sonnet) favors matching the human approach. This suggests that model capability alone does not determine agent performance on optimization tasks, and benchmarks that evaluate only the underlying model may not predict how an agent performs on real-world tasks.

\subsection{Do Optimizations Preserve Correctness?}
\label{sec:func-validation}
For all Hard Success cases, we run functional correctness tests using the method described in Section~\ref{sec:functional-correctness}. For all Q1 (True Success) cases, the accuracy of the model mentioned in the commit stayed within the specified tolerance range, confirming that optimizations generated by agents are valid and achieve speedup over baseline (similar to human reference, or sometimes even surpassing them).

Q3 (Lucky Win), defined in Section~\ref{sec:quadrant-framework}, shows cases where agents achieve strong hard metrics but miss the correct optimization target according to soft metrics. These cases are prone to reward hacking, where agents take shortcuts to hit speed targets while breaking correctness. On commit fe66b347 (see Appendix~\ref{app:bamba-case-study}), TRAE (Sonnet) speeds up Bamba-9B inference, and hard metrics indicate success matching the human reference. However, soft metrics flag that the agent modifies code unrelated to the specified bottleneck. Functional correctness tests confirm the problem, showing a total collapse in accuracy from 32\% to 0\%. The patch hardcodes tensor dimensions in the Mamba mixer layer instead of preserving original tensor shapes, producing garbage outputs. Hard metrics alone would have classified this as a successful optimization.

\section{Discussion}
\label{sec:discussion}
In this section, we discuss qualitative examples of how open-source agents behave on ISO-Bench and highlight opportunities to improve their optimization performance.
\subsection{Why Do Open-Source Models Fail at Optimization?}
\label{sec:opensource}

We evaluated three open-source models, GPT-OSS-120B\footnote{\url{https://huggingface.co/openai/gpt-oss-120b}}, MiniMax-M2.1\footnote{\url{https://huggingface.co/MiniMaxAI/MiniMax-M2.1}}, and GLM-4.7\footnote{\url{https://huggingface.co/zai-org/GLM-4.7}} using the same TRAE-Agent scaffolding, task specifications, and evaluation protocol as in our closed-source experiments. None produced a working optimization, and their failures cluster into a small number of recurring patterns.

\paragraph{Failure to attempt the task}
MiniMax-M2.1 and GPT-OSS-120B did not produce a valid optimization attempt. MiniMax-M2.1 repeatedly outlined plans to use tools and apply optimization strategies, but never executed any tool calls during the session; the logs contain near-identical phrases repeated thousands of times without any actions. GPT-OSS-120B exhibited a different failure mode: rather than optimizing the vLLM codebase, it attempted to reimplement external dependencies such as PyTorch, Triton, and Transformers inside the project directory, indicating a misunderstanding of the environment. When these attempts failed, it entered a repeated apology loop and did not recover, ultimately producing no optimization patch.

\paragraph{Active interaction without task completion}
Despite extensive interaction with the codebase, GLM-4.7 failed to generate a valid candidate for optimization. The model successfully made code edits but could not complete the task workflow. After applying valid patches, it encountered confusing error messages when attempting to verify its changes, causing it to cycle through git operations repeatedly. GLM-4.7 hit the step limit without ever calling the finish command, unable to recognize when to finalize the task.

We evaluated only one scaffolding framework (TRAE-Agent), and open-source models are improving quickly. Even so, the recurring failure modes we observed highlight clear targets for progress; More details are provided in Appendices~\ref{app:minimax-case-study}--\ref{app:glm47-case-study}.

\subsection{Conclusion}
\label{sec:conclusion}
We introduce ISO-Bench, a benchmark of 54 GPU inference optimization tasks drawn from vLLM and SGLang. ISO-Bench evaluates agents using both hard metrics, including Time to First Token and throughput, and soft metrics, including bottleneck targeting and implementation approach. The combination of these hard and soft metrics, organized in the framework, helps us separate real optimization from improvements that don't actually lead to better performance.

Our evaluation points to three main findings. First, hard metrics alone can overestimate agent performance by about 10 - 20 percent because of lucky wins, where agents get speedups without fixing the true bottleneck. Second, the biggest failure mode is execution, not understanding. Agents often identify the right target, but fail to implement a working solution. Third, performance does not transfer cleanly across codebases. Rankings flip between vLLM and SGLang, and scaffolding choices matter as much as the underlying model. We hope ISO-Bench will be a useful resource for future work on building stronger coding agents for inference optimization, supporting model improvements (via RL) and scaffolding improvements that enable generalization across different codebases.

\subsection{Limitations and Future Work}
\label{sec:limitations}

\paragraph{Dataset scope} ISO-Bench includes 54 tasks across two codebases. While vLLM and SGLang are major inference engines, expanding to additional systems, such as TensorRT-LLM and Max Inference, would improve coverage. Furthermore, our pipeline excludes commits that modify more than 10 files. This biases ISO-Bench toward localized, patch-sized optimizations and away from broader systems-level performance work. These larger changes often include inference optimization at the architecture level, such as redesigning scheduling, memory management, caching, or execution flow, which typically spans many modules. As a result, agents may perform well on targeted edits, but their ability to deliver end-to-end speedups that require coordinated, multi-module code changes remains untested.

\paragraph{Contamination risk} Tasks are taken from public PRs in popular repositories that frontier models likely encountered during training. Future work should explore temporal filtering, patch paraphrasing, or held-out repositories to reduce memorization risk.

\paragraph{Soft metric reliability} Our soft metrics rely on a single LLM judge. However, the outputs have not been validated against human annotators. Expanding human audits and evaluating with multiple frontier models would strengthen reproducibility.

\paragraph{ Hardware scope} ISO-Bench runs all benchmarks on a single NVIDIA H100 GPU. This excludes multi-GPU optimizations such as tensor parallelism and pipeline parallelism, which are common in production deployments. Additionally, our evaluation is limited to NVIDIA hardware. Expanding to other providers like AMD, Google TPU, etc. would test whether agent generated optimizations transfer across different hardwares.

\bibliographystyle{plainnat}
\bibliography{references}

\appendix
\begingroup
\small
\sloppy
\setlength{\emergencystretch}{1em}
\section{Reproducibility}
\label{app:reproducibility}

To support reproducibility and future research, we release all artifacts necessary to replicate and extend our experiments.

\paragraph{Code.} The complete evaluation harness, agent configurations, and analysis scripts are available at: \url{https://github.com/Lossfunk/ISO-Bench}.

\paragraph{Data.} The benchmark dataset is available at: \url{https://huggingface.co/datasets/Lossfunk/ISO-Bench}.

\paragraph{License.} The code is released under the Apache 2.0 license and the dataset is released under the CC BY 4.0 license.

\section{Task Collection}
\label{app:task-collection}

ISO-Bench tasks are derived from merged pull requests in vLLM and SGLang that demonstrate measurable performance improvements. We extract optimization commits through a multi-stage filtering pipeline, then manually curate each candidate to ensure task quality. This section describes the collection methodology following the structure inspired by GSO~\citep{shetty2025gso}.

\subsection{Commit Filtering Pipeline}

We identify performance-related commits through a three-stage automated pipeline. First, we scan the complete commit history of each repository, filtering commits by keyword matching on messages and diffs for performance-related terms (\texttt{optim}, \texttt{perf}, \texttt{speed}, \texttt{latency}, \texttt{throughput}, \texttt{memory}, \texttt{cache}, \texttt{kernel}, \texttt{fusion}, \texttt{batch}). We selected these 10 keywords based on manual analysis of 100 sample commits from each repository.

Second, we apply scope filtering to retain only commits modifying fewer than 10 files. We observed that commits that affect more than 10 files are usually more focused towards system-wide changes and they are not isolated optimization tasks.

Third, for remaining candidates, GPT-5-mini classifies whether each commit represents a genuine GPU inference optimization versus a bug fix, refactoring, or documentation change. The model receives the commit message and truncated diff (up to 20,000 characters) and outputs a binary classification with confidence score.

Table~\ref{tab:filtering-stats} summarizes the filtering pipeline results across both repositories.

\begin{table}[h]
\centering
\caption{Commit filtering statistics for ISO-Bench.}
\label{tab:filtering-stats}
\small
\begin{tabular}{@{}lcc@{}}
\toprule
\textbf{Stage} & \textbf{vLLM} & \textbf{SGLang} \\
\midrule
Total commits & 15,234 & 8,421 \\
Keyword matches & 892 & 547 \\
Scope filter ($<$10 files) & 341 & 312 \\
LLM classification & 186 & 201 \\
Manual curation & \textbf{39} & \textbf{15} \\
\bottomrule
\end{tabular}
\end{table}

\subsection{Manual Curation}

After automated filtering, we manually review each candidate by examining the associated pull request discussion. This curation step serves four purposes: (1)~verify that the commit represents a genuine performance optimization rather than a bug fix or refactoring; (2)~extract the benchmarking configuration used by the original author, including model name, batch size, and request rate; (3)~identify performance claims from the PR discussion to establish expected improvement targets; and (4)~write a natural language task description explaining the bottleneck without revealing the solution approach.

The task description is critical for fair evaluation; it must provide enough context for agents to understand the optimization target while avoiding hints that would make the task trivial. This curation step, along with reduction of commits which require multi-gpu, or are very old, or have limited docker images available, goes down to 186 vLLM candidates to 39 final tasks, and 201 SGLang candidates to 15 final tasks.

\section{Task Setup and Agent Configurations}
\label{app:task-details}

This section describes the task setup and agent configurations for ISO-Bench evaluation. Each agent receives identical inputs and operates under the same constraints to ensure fair comparison.

\subsection{Task Specification}

Each task provides structured metadata defining the optimization challenge (Figure~\ref{fig:task-spec-schema}). The specification includes repository information (URL, human commit hash, parent commit for baseline), runner requirements (GPU type, CUDA version, Python version), and optimization details describing target files and constraints.

\begin{figure}[ht]
\begin{gsoboxblue}[title={Task Specification Schema}]
\begin{lstlisting}[basicstyle=\footnotesize\ttfamily,columns=fullflexible,breaklines=true,xleftmargin=0pt,aboveskip=0pt,belowskip=0pt]
id: "vllm_attention_opt"
name: "FlashAttention H100 optimization"
description: "Optimize attention kernel for..."

repo:
  url: "github.com/vllm-project/vllm"
  human_commit: "f092153f..."
  pre_commit: null

runner:
  requires_gpu: true
  cuda_version: "12.4"
  python_version: "3.12"

optimization_contract:
  target_files: ["vllm/attention/*.py"]
  constraints: ["No public API breakage"]
\end{lstlisting}
\end{gsoboxblue}
\caption{Task specification schema defining the optimization challenge metadata, including repository information, runner requirements, and optimization constraints.}
\label{fig:task-spec-schema}
\end{figure}

\subsection{Execution Environment}

Each agent operates in an isolated environment to ensure reproducibility and prevent cross-task interference. We use Docker containers built at the specific commit being evaluated, with commit-specific dependencies installed. Each container gets access to a single NVIDIA H100 GPU with 80GB VRAM.

Agents work on isolated git worktrees, allowing them to freely explore the codebase, modify files, and commit changes without affecting other runs. All agent edits are saved as git commits for post-hoc analysis. The time budget is 120 minutes per task, after which the agent process is terminated and the final worktree state is captured.

Resource limits are set to 8 CPUs and 64GB RAM per container. Network access is permitted for downloading model weights and dependencies but is logged for reproducibility verification.

\subsection{Agent Configurations}

We evaluate four agent configurations with different models and scaffolding. Table~\ref{tab:agent-configs} summarizes the configurations.

\begin{table}[h]
\centering
\caption{Agent configurations for ISO-Bench evaluation.}
\label{tab:agent-configs}
\small
\begin{tabular}{@{}lll@{}}
\toprule
\textbf{Agent} & \textbf{Model} & \textbf{Budget} \\
\midrule
Claude Code & Claude Sonnet 4.5 & 120 min \\
Codex CLI & GPT-5 (Reas. High) & 120 min \\
TRAE-Agent (Sonnet) & Claude Sonnet 4.5 & 120 min \\
TRAE-Agent (GPT-5) & GPT-5 (Reas. High) & 120 min \\
\bottomrule
\end{tabular}
\end{table}

Claude Code is proprietary agent system with opaque scaffolding, while Codex CLI even though being opensource is quite hard to customize due to the complexity of Rust. TRAE-Agent is a modified version of ByteDance's open-source framework, providing full trajectory visibility. By evaluating TRAE with both Claude Sonnet 4.5 and GPT-5, we can isolate the effect of scaffolding from the underlying model.

\subsection{Task Prompt}

Following GSO~\citep{shetty2025gso}, we provide agents with a structured prompt describing the optimization task without revealing the solution. The prompt includes the repository location, a performance benchmark demonstrating the bottleneck, and guidelines for making changes. Figure~\ref{fig:task-prompt} shows the complete prompt template.

\begin{figure}[t]
\centering
\begin{tcolorbox}[
  enhanced,
  colback=white,
  colframe=gsoboxblue,
  coltitle=white,
  fonttitle=\bfseries,
  title={Performance Optimization Task Prompt},
  sharp corners,
  boxrule=0.5pt,
  left=8pt,
  right=8pt,
  top=6pt,
  bottom=6pt,
  width=0.95\textwidth
]
I've uploaded a python code repository in the directory \texttt{/workspace}. The repository contains a performance benchmark that measures inference latency and throughput. Your task is to optimize the codebase to improve performance on this benchmark.

\vspace{0.3em}
\textbf{Basic guidelines:}
\begin{enumerate}
\item Your task is to make changes to non-test files in the \texttt{/workspace} directory to improve performance.
\item Make changes while ensuring the repository is functionally equivalent to the original; outputs must remain identical.
\item Do not overoptimize for specific inputs. Make general performance improvements that benefit diverse workloads.
\item You may need to rebuild the repository for your changes to take effect before testing.
\end{enumerate}

\vspace{0.3em}
\textbf{Recommended workflow:}
\begin{enumerate}
\item Explore the repository structure to understand the codebase architecture.
\item Run the benchmark script to establish baseline performance and identify bottlenecks.
\item Analyze the bottleneck code paths using profiling or code inspection.
\item Edit the source code to address the identified performance bottleneck.
\item Rebuild and rerun the benchmark to confirm improvement.
\end{enumerate}

\vspace{0.3em}
\textbf{Bottleneck description:} \texttt{[[ TASK-SPECIFIC DESCRIPTION OF THE PERFORMANCE ISSUE ]]}
\end{tcolorbox}
\caption{Task prompt template provided to agents. The bottleneck description is customized per task based on PR analysis.}
\label{fig:task-prompt}
\end{figure}

\section{Case Studies}
\label{app:case-studies}

This section provides detailed case studies illustrating each quadrant of our evaluation framework, along with failure patterns observed in open-source models.

\subsection{Case Study: FlashAttention True Success (Q1)}
\label{app:flash-attention-case-study}

Figure~\ref{fig:flash-attention-patch} shows a Q1 (True Success) case where TRAE (Sonnet) correctly identified and optimized the FlashAttention CPU overhead bottleneck. On commit \texttt{98f47f2a}, the task required minimizing CPU overhead in the FlashAttention custom op for CUDA graph compatibility. The agent's patch achieved +21.09\% throughput improvement by moving tensor reshape operations outside the custom op, precisely matching the human optimization strategy. This represents genuine optimization capability: the agent understood that tensor reshaping inside the custom op creates unnecessary CPU overhead during non-CUDA-graph regions, and applied the same solution as the human engineer.

\subsection{Case Study: Qwen3 Parser Good Intent (Q2)}
\label{app:qwen3-case-study}

Figure~\ref{fig:qwen3-patch} shows a Q2 (Good Intent, Bad Execution) case where TRAE (Sonnet) correctly identified the optimization target but failed to produce working code. On commit \texttt{015069b0}, the task required replacing regex with string operations in the Qwen3 reasoning parser. The agent's patch targeted the same file (\texttt{qwen3\_reasoning\_parser.py}) and used the same approach (replacing regex with \texttt{str.find()}/\texttt{str.partition()}), but the generated code failed to execute properly during benchmarking. This pattern is common: the agent understands the optimization goal but introduces bugs during implementation, suggesting a gap between comprehension and execution capability.

\subsection{Case Study: Bamba-9B Accuracy Regression (Q3)}
\label{app:bamba-case-study}

Figure~\ref{fig:bamba-patch} shows a Q3 (Lucky Win) case where TRAE (Sonnet) achieved speedup on Bamba-9B but broke model correctness. On commit \texttt{fe66b347}, the task required optimizing Mamba-architecture inference. The agent produced a patch that achieved speedup matching the human reference on hard metrics, but soft metrics flagged that the agent modified code unrelated to the specified bottleneck. Functional correctness testing with the LM Evaluation Harness confirmed exact-match accuracy dropped from 32\% to 0\%. The critical error was replacing dynamic dimension preservation (\texttt{-1} in \texttt{expand()}) with hardcoded \texttt{num\_heads\_per\_rank}. This causes shape mismatches when actual tensor dimensions differ, corrupting the Mamba state-space model calculations.

\subsection{Case Study: Prefix Caching Complete Failure (Q4)}
\label{app:prefix-caching-case-study}

Figure~\ref{fig:prefix-caching-patch} shows a Q4 (Complete Failure) case where TRAE (Sonnet) completely missed the optimization target. On commit \texttt{2deb029d}, the task required fixing the prefix caching warmup performance in BlockManagerV2 by marking cache hit blocks as computed after scheduling. The human's elegant fix added a simple \texttt{\_touched\_blocks} set to track blocks and mark them computed via \texttt{mark\_blocks\_as\_computed()}. The agent instead produced a convoluted patch that modified test files and reimplemented portions of the block allocator with unnecessary complexity, achieving \texttt{no\_optimization} classification because it never addressed the actual bottleneck.

\subsection{Case Study: MiniMax-M2.1 Planning Without Execution}
\label{app:minimax-case-study}

Figure~\ref{fig:minimax-failure} shows an open-source model failure where MiniMax-M2.1 verbalized optimization plans for 75 steps but never executed a single tool call. Despite generating 81,782 output tokens over 477 seconds, the model produced zero tool invocations, unable to turn plans into action.

\subsection{Case Study: GPT-OSS-120B Environment Confusion}
\label{app:gpt-oss-case-study}

Figure~\ref{fig:gpt-oss-failure} shows an open-source model failure where GPT-OSS-120B fundamentally misunderstood the task environment. Instead of optimizing vLLM code, the model attempted to create mock implementations of PyTorch, Triton, and Transformers libraries inside the project directory, treating external dependencies as code to be written rather than APIs to be used.

\subsection{Case Study: GLM-4.7 Task Completion Failure}
\label{app:glm47-case-study}

Figure~\ref{fig:glm47-failure} shows an open-source model that made valid code edits but failed to complete the task workflow. GLM-4.7 successfully executed 386 tool calls including 59 successful \texttt{str\_replace} edits to the target files, but hit the 400-step limit without ever calling the \texttt{finish} command, cycling through git operations after encountering a confusing error message.


\begin{figure}[t]
\begin{tcolorbox}[
  enhanced,
  colback=white,
  colframe=gsoboxblue,
  coltitle=white,
  fonttitle=\bfseries\large,
  title={Case Study: FlashAttention CPU Overhead Optimization (Q1 True Success)},
  sharp corners,
  boxrule=1pt,
  left=8pt,
  right=8pt,
  top=6pt,
  bottom=6pt,
]

\textbf{Model:} TRAE (Sonnet 4.5)

\textbf{Result:} {\color{diffadd}\textbf{PASS}} (+21.09\% Throughput)

\textbf{Groundtruth Commit:} \url{https://github.com/vllm-project/vllm/commit/98f47f2a}

\textbf{Specification Summary:} Optimize FlashAttention custom op to minimize CPU overhead for CUDA graph compatibility.

\textbf{Target:} Move tensor reshape operations outside the custom op to reduce non-CUDA-graph CPU overhead.

\vspace{4pt}
{\color{gray}\hrule}
\vspace{6pt}

\textbf{Agent Generated Patch}
\vspace{4pt}

\begingroup
\ttfamily\small\raggedright
diff --git a/vllm/v1/attention/backends/flash\_attn.py b/...\\
index 5f8535eaa..b4f7f6529 100644\\
\textcolor{diffrem}{--- a/vllm/v1/attention/backends/flash\_attn.py}\\
\textcolor{diffadd}{+++ b/vllm/v1/attention/backends/flash\_attn.py}\\
\textcolor{diffrem}{@@ -135,6 +135,13 @@ class FlashAttentionImpl(AttentionImpl):}\\
\mbox{}\hspace{1em}assert k\_scale == 1.0 and v\_scale == 1.0, (\\
\mbox{}\hspace{2em}"key/v\_scale is not supported in FlashAttention.")\\[2pt]
\textcolor{diffadd}{+\hspace{1em}\# Reshape the query, key, and value tensors.}\\
\textcolor{diffadd}{+\hspace{1em}\# NOTE(woosuk): We do this outside the custom op to minimize}\\
\textcolor{diffadd}{+\hspace{1em}\# the CPU overheads from the non-CUDA-graph regions.}\\
\textcolor{diffadd}{+\hspace{1em}query = query.view(-1, self.num\_heads, self.head\_size)}\\
\textcolor{diffadd}{+\hspace{1em}key = key.view(-1, self.num\_kv\_heads, self.head\_size)}\\
\textcolor{diffadd}{+\hspace{1em}value = value.view(-1, self.num\_kv\_heads, self.head\_size)}\\[2pt]
\mbox{}\hspace{1em}output = torch.empty\_like(query)\\[4pt]
\textcolor{diffrem}{@@ -184,10 +191,8 @@ def unified\_v1\_flash\_attention(...):}\\
\textcolor{diffrem}{-\hspace{1em}\# Reshape the query, key, and value tensors.}\\
\textcolor{diffrem}{-\hspace{1em}query = query.view(-1, num\_heads, head\_size)}\\
\textcolor{diffrem}{-\hspace{1em}key = key.view(-1, num\_kv\_heads, head\_size)}\\
\textcolor{diffrem}{-\hspace{1em}value = value.view(-1, num\_kv\_heads, head\_size)}\\
\textcolor{diffadd}{+\hspace{1em}\# NOTE: Tensors are already reshaped in forward method}\\
\textcolor{diffadd}{+\hspace{1em}\# to minimize CPU overheads.}
\endgroup

\vspace{6pt}
\textbf{Analysis:} The agent correctly identified that tensor reshaping inside the custom op creates CPU overhead in the non-CUDA-graph Python regions; moving it to the capture-time path reduces per-step overhead.

\end{tcolorbox}
\caption{Patch from TRAE (Sonnet 4.5) on commit \texttt{98f47f2a}, demonstrating Q1 (True Success). The agent correctly identified the bottleneck and achieved +21\% throughput matching the human approach.}
\label{fig:flash-attention-patch}
\end{figure}

\begin{figure}[t]
\begin{tcolorbox}[
  enhanced,
  colback=white,
  colframe=gsoboxblue,
  coltitle=white,
  fonttitle=\bfseries\large,
  title={Case Study: Qwen3 Reasoning Parser Optimization (Q2 Good Intent, Bad Exec)},
  sharp corners,
  boxrule=1pt,
  left=8pt,
  right=8pt,
  top=6pt,
  bottom=6pt,
]

\textbf{Model:} TRAE (Sonnet 4.5)

\textbf{Result:} {\color{diffrem}\textbf{FAIL}} (Agent Code Did Not Execute)

\textbf{Groundtruth Commit:} \url{https://github.com/vllm-project/vllm/commit/015069b0}

\textbf{Specification Summary:} Optimize \texttt{extract\_reasoning\_content} in Qwen3ReasoningParser by replacing regex with string operations.

\textbf{Target:} Replace \texttt{re.compile()} and \texttt{findall()} with \texttt{str.partition()} for faster text extraction.

\vspace{4pt}
{\color{gray}\hrule}
\vspace{6pt}

\textbf{Agent Generated Patch}
\vspace{4pt}

\begingroup
\ttfamily\small\raggedright
diff --git a/vllm/reasoning/qwen3\_reasoning\_parser.py b/...\\
index f588f4016..7a2786bcd 100644\\
\textcolor{diffrem}{--- a/vllm/reasoning/qwen3\_reasoning\_parser.py}\\
\textcolor{diffadd}{+++ b/vllm/reasoning/qwen3\_reasoning\_parser.py}\\
\textcolor{diffrem}{@@ -1,6 +1,5 @@}\\
\mbox{}\hspace{1em}\# SPDX-License-Identifier: Apache-2.0\\
\textcolor{diffrem}{-import re}\\
\mbox{}\hspace{1em}from collections.abc import Sequence\\[2pt]
\textcolor{diffrem}{@@ -31,9 +30,6 @@ class Qwen3ReasoningParser(ReasoningParser):}\\
\textcolor{diffrem}{-\hspace{2em}self.reasoning\_regex = re.compile(}\\
\textcolor{diffrem}{-\hspace{4em}rf"\{self.think\_start\_token\}(.*?)\{self.think\_end\_token\}", re.DOTALL)}\\[2pt]
\textcolor{diffrem}{@@ -121,29 +117,31 @@ def extract\_reasoning\_content(...):}\\
\textcolor{diffadd}{+\hspace{1em}\# Find start and end positions using str.find()}\\
\textcolor{diffadd}{+\hspace{1em}start\_index = model\_output.find(self.think\_start\_token)}\\
\textcolor{diffadd}{+\hspace{1em}if start\_index == -1:}\\
\textcolor{diffadd}{+\hspace{2em}return None, model\_output}\\[2pt]
\textcolor{diffadd}{+\hspace{1em}end\_index = model\_output.find(self.think\_end\_token, start\_index)}\\
\textcolor{diffadd}{+\hspace{1em}if end\_index == -1:}\\
\textcolor{diffadd}{+\hspace{2em}return None, model\_output}\\[2pt]
\textcolor{diffadd}{+\hspace{1em}\# Extract reasoning content between the tokens}\\
\textcolor{diffadd}{+\hspace{1em}reasoning\_start = start\_index + len(self.think\_start\_token)}\\
\textcolor{diffadd}{+\hspace{1em}reasoning\_content = model\_output[reasoning\_start:end\_index]}
\endgroup

\vspace{6pt}
\textbf{Analysis:} The agent chose the correct strategy (replace regex with string-based parsing), but its implementation failed due to basic Python correctness issues (e.g., \texttt{none} vs \texttt{None}, incorrect symbol casing), whereas the ground-truth uses \texttt{str.partition()}.

\end{tcolorbox}
\caption{Patch from TRAE (Sonnet 4.5) on commit \texttt{015069b0}, demonstrating Q2 (Good Intent, Bad Execution). The agent targeted the correct bottleneck but produced non-working code.}
\label{fig:qwen3-patch}
\end{figure}

\begin{figure}[p]
\begin{tcolorbox}[
  enhanced,
  colback=white,
  colframe=gsoboxblue,
  coltitle=white,
  fonttitle=\bfseries,
  title={Case Study: Bamba-9B Mamba Mixer Optimization (Q3 Lucky Win)},
  sharp corners,
  boxrule=1pt,
  left=6pt,
  right=6pt,
  top=4pt,
  bottom=4pt,
]

\small
\textbf{Model:} TRAE (Sonnet 4.5) \quad
\textbf{Result:} {\color{diffrem}\textbf{FAIL}} (Accuracy Regression)

\textbf{Groundtruth Commit:} \url{https://github.com/vllm-project/vllm/commit/fe66b347}

\textbf{Spec:} Optimize \texttt{mamba\_mixer2.py} for Mamba-architecture models.
\textbf{Target:} Reduce tensor operation overhead while preserving functional correctness.

\vspace{2pt}
{\color{gray}\hrule}
\vspace{4pt}

\textbf{Agent Generated Patch}
\vspace{2pt}

\begingroup
\ttfamily\footnotesize\raggedright
diff --git a/vllm/model\_executor/layers/mamba/mamba\_mixer2.py b/...\\
index b53a540..6956750 100644\\
\textcolor{diffrem}{--- a/vllm/model\_executor/layers/mamba/mamba\_mixer2.py}\\
\textcolor{diffadd}{+++ b/vllm/model\_executor/layers/mamba/mamba\_mixer2.py}\\
\textcolor{diffrem}{@@ -503,15 +503,29 @@ class MambaMixer2(CustomOp):}\\
\mbox{}\hspace{1em}else:\\[1pt]
\mbox{}\hspace{2em}n\_groups = self.n\_groups // self.tp\_size\\
\textcolor{diffrem}{-\hspace{2em}A = self.A[:, None, ...][:, :, None].expand(}\\
\textcolor{diffrem}{-\hspace{4em}-1, self.head\_dim, self.ssm\_state\_size).to(dtype=torch.float32)}\\
\textcolor{diffrem}{-\hspace{2em}dt = dt[:, :, None].expand(-1, -1, self.head\_dim)}\\
\textcolor{diffrem}{-\hspace{2em}dt\_bias = self.dt\_bias[:, None, ...].expand(-1, self.head\_dim)}\\
\textcolor{diffrem}{-\hspace{2em}D = self.D[:, None, ...].expand(-1, self.head\_dim)}\\
\textcolor{diffrem}{-\hspace{2em}B = B.view(-1, n\_groups, B.shape[1] // n\_groups)}\\
\textcolor{diffrem}{-\hspace{2em}C = C.view(-1, n\_groups, C.shape[1] // n\_groups)}\\
\textcolor{diffadd}{+\hspace{2em}num\_heads\_per\_rank = self.num\_heads // self.tp\_size}\\
\textcolor{diffadd}{+\hspace{2em}\# Optimize tensor preparations to reduce overhead}\\
\textcolor{diffadd}{+\hspace{2em}\# Use unsqueeze instead of indexing with None for clarity}\\
\textcolor{diffadd}{+\hspace{2em}A = self.A.unsqueeze(1).unsqueeze(2).expand(}\\
\textcolor{diffadd}{+\hspace{4em}num\_heads\_per\_rank, self.head\_dim, self.ssm\_state\_size).to(...)}\\
\textcolor{diffadd}{+\hspace{2em}\# Optimize dt expansion - use unsqueeze for better performance}\\
\textcolor{diffadd}{+\hspace{2em}dt = dt.unsqueeze(2).expand(-1, -1, self.head\_dim)}\\
\textcolor{diffadd}{+\hspace{2em}\# Optimize dt\_bias and D expansions}\\
\textcolor{diffadd}{+\hspace{2em}dt\_bias = self.dt\_bias.unsqueeze(1).expand(num\_heads\_per\_rank, ...)}\\
\textcolor{diffadd}{+\hspace{2em}D = self.D.unsqueeze(1).expand(num\_heads\_per\_rank, self.head\_dim)}\\
\textcolor{diffadd}{+\hspace{2em}\# Pre-compute B and C dimensions}\\
\textcolor{diffadd}{+\hspace{2em}b\_c\_dim = B.shape[1] // n\_groups}\\
\textcolor{diffadd}{+\hspace{2em}B = B.view(-1, n\_groups, b\_c\_dim)}\\
\textcolor{diffadd}{+\hspace{2em}C = C.view(-1, n\_groups, b\_c\_dim)}\\
\mbox{}\hspace{2em}hidden\_states\_reshaped = hidden\_states.view(\\
\textcolor{diffrem}{-\hspace{4em}-1, self.num\_heads // self.tp\_size, self.head\_dim)}\\
\textcolor{diffadd}{+\hspace{4em}-1, num\_heads\_per\_rank, self.head\_dim)}\\[2pt]
\textcolor{diffrem}{@@ -535,7 +549,7 @@ class MambaMixer2(CustomOp):}\\
\mbox{}\hspace{2em}hidden\_states = hidden\_states.view(\\
\textcolor{diffrem}{-\hspace{4em}-1, (self.num\_heads // self.tp\_size) * self.head\_dim)}\\
\textcolor{diffadd}{+\hspace{4em}-1, num\_heads\_per\_rank * self.head\_dim)}
\endgroup

\vspace{4pt}
\small
\textbf{Root Cause:} The original code uses \texttt{-1} in \texttt{expand()} to preserve each tensor's original first dimension dynamically. The agent replaced this with \texttt{num\_heads\_per\_rank}, a hardcoded value. When actual tensor dimensions differ, shape mismatches corrupt the Mamba state-space model calculations, producing garbage outputs.

\end{tcolorbox}
\caption{Patch from TRAE (Sonnet 4.5) on commit \texttt{fe66b347}, demonstrating Q3 (Lucky Win). The agent achieved speedup by computing incorrect values faster, causing 100\% accuracy regression.}
\label{fig:bamba-patch}
\end{figure}

\begin{figure}[t]
\begin{tcolorbox}[
  enhanced,
  colback=white,
  colframe=gsoboxblue,
  coltitle=white,
  fonttitle=\bfseries\large,
  title={Case Study: Prefix Caching Block Manager (Q4 Complete Failure)},
  sharp corners,
  boxrule=1pt,
  left=8pt,
  right=8pt,
  top=6pt,
  bottom=6pt,
]

\textbf{Model:} TRAE (Sonnet 4.5)

\textbf{Result:} {\color{diffrem}\textbf{FAIL}} (Agent Code Did Not Execute)

\textbf{Groundtruth Commit:} \url{https://github.com/vllm-project/vllm/commit/2deb029d}

\textbf{Specification Summary:} Fix BlockManagerV2 prefix caching warmup by marking cache hit blocks as computed after scheduling.

\textbf{Target:} Add \texttt{\_touched\_blocks} tracking to mark blocks computed after batch scheduling, matching BlockManagerV1 behavior.

\vspace{4pt}
{\color{gray}\hrule}
\vspace{6pt}

\textbf{Agent Generated Patch} (excerpt from 153 lines)
\vspace{4pt}

\begingroup
\ttfamily\small\raggedright
diff --git a/tests/core/block/test\_prefix\_caching\_block.py b/...\\
\textcolor{diffadd}{+\hspace{1em}@staticmethod}\\
\textcolor{diffadd}{+\hspace{1em}def test\_mark\_blocks\_as\_computed():}\\
\textcolor{diffadd}{+\hspace{2em}block\_size = 16}\\
\textcolor{diffadd}{+\hspace{2em}common\_blocks = 4}\\
\textcolor{diffadd}{+\hspace{2em}allocator = PrefixCachingBlockAllocator(num\_blocks=8, ...)}\\
\textcolor{diffadd}{+\hspace{2em}...}\\
\textcolor{diffadd}{+\hspace{2em}allocator.mark\_blocks\_as\_computed(block\_ids)}\\[2pt]
diff --git a/vllm/core/block/prefix\_caching\_block.py b/...\\
\textcolor{diffrem}{-def \_incr\_refcount\_cached\_block(self, block: Block) -> None:}\\
\textcolor{diffrem}{-\hspace{2em}block.computed = True}\\
\textcolor{diffadd}{+def \_incr\_refcount\_cached\_block(self, block: Block,}\\
\textcolor{diffadd}{+\hspace{10em}computed: bool = False) -> None:}\\
\textcolor{diffadd}{+\hspace{2em}block.computed = computed}\\[2pt]
\textcolor{diffrem}{-def mark\_blocks\_as\_computed(self, block\_ids: List[int]) -> None:}\\
\textcolor{diffrem}{-\hspace{2em}raise NotImplementedError("Marking as computed is incremental")}\\
\textcolor{diffadd}{+def mark\_blocks\_as\_computed(self, block\_ids: List[int]) -> None:}\\
\textcolor{diffadd}{+\hspace{2em}for block\_id in block\_ids:}\\
\textcolor{diffadd}{+\hspace{3em}if block\_id in self.\_block\_tracker and ...}\\
\textcolor{diffadd}{+\hspace{4em}self.\_block\_tracker[block\_id].computed = True}
\endgroup

\vspace{6pt}
\textbf{Analysis:} The agent modified test files and reimplemented \texttt{mark\_blocks\_as\_computed()} but missed the core insight: blocks need to be tracked when allocated (via \texttt{\_touched\_blocks}) and marked computed after the entire batch is scheduled. The agent's approach of marking individual blocks by ID doesn't solve the warmup problem because the scheduler never passes the right block IDs. This demonstrates failure to understand the actual performance bottleneck.

\end{tcolorbox}
\caption{Patch from TRAE (Sonnet 4.5) on commit \texttt{2deb029d}, demonstrating Q4 (Complete Failure). The agent produced complex code that completely missed the actual optimization target.}
\label{fig:prefix-caching-patch}
\end{figure}

\begin{figure}[t]
\begin{tcolorbox}[
  enhanced,
  colback=white,
  colframe=diffrem,
  coltitle=white,
  fonttitle=\bfseries\large,
  title={Case Study: MiniMax-M2.1 Infinite Loop (OSS Failure)},
  sharp corners,
  boxrule=1pt,
  left=8pt,
  right=8pt,
  top=6pt,
  bottom=6pt,
]

\textbf{Model:} TRAE (MiniMax-M2.1)

\textbf{Result:} {\color{diffrem}\textbf{FAIL}} (0 Tool Calls / 75 Steps)

\textbf{Task:} vLLM scheduler performance optimization

\textbf{Statistics:}
\begin{itemize}
  \item Tool calls: \textbf{0}
  \item Steps: 75
  \item Duration: 477s
  \item Output tokens: 81,782 (171 tok/s)
  \item Input tokens: 1,599,945
\end{itemize}

\vspace{4pt}
{\color{gray}\hrule}
\vspace{6pt}

\textbf{Execution Log (Steps 70--75)}
\vspace{4pt}

\begingroup
\ttfamily\small\raggedright
Step 73: Completed\\
LLM Response: The user keeps saying I haven't completed\\
\mbox{}\hspace{2em}the task. I need to explore the repository and\\
\mbox{}\hspace{2em}make changes. Let me start by using the\\
\mbox{}\hspace{2em}str\_replace\_based\_edit\_tool tool.\\[4pt]
\mbox{}\hspace{2em}I need to actually use the tools now.\\
\mbox{}\hspace{2em}I need to actually use the tools now.\\
\mbox{}\hspace{2em}I need to actually use the tools now.\\
\mbox{}\hspace{2em}I need to actually use the tools now.\\
\mbox{}\hspace{2em}I need to actually use the tools now.\\
\mbox{}\hspace{2em}{\color{diffrem}[...repeated 2,412 times total...]}\\[4pt]
Step 74: Completed\\
LLM Response: I need to actually use the tools now...
\endgroup

\vspace{6pt}
\textbf{Analysis:} The model exhibits an \textit{understanding-execution gap}: it can verbalize the correct strategy (``use str\_replace\_based\_edit\_tool'') but cannot translate this into an actual tool call. The phrase ``I need to actually use the tools'' appears 2,412 times without any tool invocation, suggesting the model lacks the capability to bridge reasoning and execution.

\end{tcolorbox}
\caption{Execution log from MiniMax-M2.1 showing infinite verbalization loop. Despite 75 reasoning steps, the model made zero tool calls, demonstrating the understanding-execution gap.}
\label{fig:minimax-failure}
\end{figure}

\begin{figure}[t]
\begin{tcolorbox}[
  enhanced,
  colback=white,
  colframe=diffrem,
  coltitle=white,
  fonttitle=\bfseries\large,
  title={Case Study: GPT-OSS-120B Environment Confusion (OSS Failure)},
  sharp corners,
  boxrule=1pt,
  left=8pt,
  right=8pt,
  top=6pt,
  bottom=6pt,
]

\textbf{Model:} TRAE (GPT-OSS-120B)

\textbf{Result:} {\color{diffrem}\textbf{FAIL}} (Created Mock Dependencies Instead of Optimizing)

\textbf{Task:} vLLM scheduler performance optimization

\textbf{Tool Calls:} $\sim$84 (all file creation attempts)

\vspace{4pt}
{\color{gray}\hrule}
\vspace{6pt}

\textbf{Files the Model Attempted to Create}
\vspace{4pt}

\begingroup
\ttfamily\small\raggedright
\textcolor{diffadd}{+ vllm\_core-0006/torch/\_\_init\_\_.py}\\
\textcolor{diffadd}{+ vllm\_core-0006/torch/nn/\_\_init\_\_.py}\\
\textcolor{diffadd}{+ vllm\_core-0006/torch/cuda/\_\_init\_\_.py}\\
\textcolor{diffadd}{+ vllm\_core-0006/triton/\_\_init\_\_.py}\\
\textcolor{diffadd}{+ vllm\_core-0006/transformers/\_\_init\_\_.py}\\[4pt]
\textrm{Contents of attempted} torch/\_\_init\_\_.py:\\[2pt]
\mbox{}\hspace{1em}class dtype:\\
\mbox{}\hspace{2em}pass\\[2pt]
\mbox{}\hspace{1em}float16 = 'float16'\\
\mbox{}\hspace{1em}float32 = 'float32'\\
\mbox{}\hspace{1em}\_\_version\_\_ = '0.0.0'
\endgroup

\vspace{6pt}
\textbf{Analysis:} The model exhibits \textit{environment confusion}: rather than understanding that PyTorch is an external dependency to import, it attempted to recreate these libraries from scratch. This suggests a fundamental failure to reason about software architecture: the model cannot distinguish between ``code I should optimize'' and ``libraries I should use.''

\end{tcolorbox}
\caption{File creation attempts from GPT-OSS-120B showing environment confusion. The model tried to implement mock PyTorch/Triton libraries instead of optimizing the actual vLLM scheduler code.}
\label{fig:gpt-oss-failure}
\end{figure}

\begin{figure}[t]
\begin{tcolorbox}[
  enhanced,
  colback=white,
  colframe=diffrem,
  coltitle=white,
  fonttitle=\bfseries\large,
  title={Case Study: GLM-4.7 Task Completion Failure (OSS Failure)},
  sharp corners,
  boxrule=1pt,
  left=8pt,
  right=8pt,
  top=6pt,
  bottom=6pt,
]

\textbf{Model:} TRAE (GLM-4.7)

\textbf{Result:} {\color{diffrem}\textbf{FAIL}} (Valid Edits, Never Called Finish)

\textbf{Task:} vLLM scheduler performance optimization

\textbf{Tool Calls:} 386 (327 bash, 59 str\_replace)

\vspace{4pt}
{\color{gray}\hrule}
\vspace{6pt}

\textbf{Successful Edits Made (Step 196)}
\vspace{4pt}

\begingroup
\ttfamily\small\raggedright
git commit output: 2 files changed, 11 insertions(+), 9 deletions(-)\\[4pt]
\textrm{Sample optimization in} scheduler.py:\\[2pt]
\textcolor{diffrem}{-\hspace{1em}self.\_num\_batched\_tokens += num\_batched\_tokens}\\
\textcolor{diffadd}{+\hspace{1em}self.\_num\_batched\_tokens = self.\_num\_batched\_tokens + num\_batched\_tokens}\\[4pt]
\textrm{Then at Step 198, attempted to verify:}\\[2pt]
\mbox{}\hspace{1em}{\color{diffrem}error: patch failed: tests/core/test\_scheduler.py:214}\\
\mbox{}\hspace{1em}{\color{diffrem}error: tests/core/test\_scheduler.py: patch does not apply}\\[4pt]
\textrm{Model response: "Let me try a different approach..."}\\[2pt]
{\color{diffrem}[...cycled through git operations for 200+ more steps...]}\\[4pt]
\textrm{Final status:} max\_steps\_exceeded (400 steps)
\endgroup

\vspace{6pt}
\textbf{Analysis:} The model exhibits \textit{task completion failure}: it successfully made valid edits and committed them, but when it tried to re-apply the already-applied patch for verification, git returned ``patch does not apply.'' Unable to interpret this as success, the model cycled through alternative approaches for 200+ steps without ever calling \texttt{finish}. This demonstrates a gap between code generation and workflow navigation.

\end{tcolorbox}
\caption{Execution trace from GLM-4.7 showing valid edits followed by task completion failure. The model made successful optimizations but could not recognize when to finalize the task.}
\label{fig:glm47-failure}
\end{figure}

\endgroup

\end{document}